%% file: main-sigconf.tex
\documentclass[sigconf]{acmart}

\usepackage{hyperref}

\usepackage{graphicx}
\usepackage{comment}
\usepackage{amsmath} 
\usepackage{color}

\usepackage[misc,geometry]{ifsym}
\usepackage{bm}
\usepackage{booktabs}
\usepackage{multirow}
\usepackage{times}
\usepackage{epsfig}
\usepackage{subfigure}
\usepackage{diagbox}
\usepackage{makecell}
\usepackage{colortbl}
\usepackage{threeparttable}

\usepackage{algorithm}
\usepackage{bm}

\usepackage{algpseudocode}
\usepackage{setspace}
\input{math_commands.tex}

\newcommand{\tablestyle}[2]{\setlength{\tabcolsep}{#1}\renewcommand{\arraystretch}{#2}\centering\footnotesize}
\newlength\savewidth

\AtBeginDocument{%
  \providecommand\BibTeX{{%
    \normalfont B\kern-0.5em{\scshape i\kern-0.25em b}\kern-0.8em\TeX}}}

\copyrightyear{2021}
\acmYear{2021}
\setcopyright{acmcopyright}\acmConference[MM '21]{Proceedings of the 29th ACM
International Conference on Multimedia}{October 20--24, 2021}{Virtual Event, China}
\acmBooktitle{Proceedings of the 29th ACM International Conference on Multimedia
(MM '21), October 20--24, 2021, Virtual Event, China}
\acmPrice{15.00}
\acmDOI{10.1145/3474085.3475414}
\acmISBN{978-1-4503-8651-7/21/10}

\begin{document}

\title{Elastic Tactile Simulation Towards Tactile-Visual Perception}

\author{Yikai Wang$^1$, Wenbing Huang$^{1}$,  Bin Fang$^{1}$,  Fuchun Sun$^{1}\,$\textsuperscript{\Letter}, Chang Li$^{2}$}
\affiliation{%
  \institution{$^{1}$Beijing National Research Center for Information Science and Technology$\,$(BNRist),\\ State Key Lab on Intelligent Technology and Systems,\\Department of Computer Science and Technology, Tsinghua University$\;$ $^{2}$JD Explore Academy}
  \country{\texttt{\normalsize wangyk17@mails.tsinghua.edu.cn, hwenbing@126.com,\\\{fangbin, fcsun\}@mail.tsinghua.edu.cn, lichang93@jd.com}
}}

\renewcommand{\shortauthors}{Y. Wang, W. Huang, B. Fang, F. Sun, C. Li}

\begin{abstract}
Tactile sensing plays an important role in robotic perception and manipulation tasks. To overcome the real-world limitations of data collection, simulating tactile response in a virtual environment comes as a desirable direction of robotic research. In this paper, we propose Elastic Interaction of Particles (EIP) for tactile simulation, which is capable of reflecting the elastic property of the tactile sensor as well as characterizing the fine-grained physical interaction during contact. Specifically, EIP models the tactile sensor as a group of coordinated particles, and the elastic property is applied to regulate the deformation of particles during contact. With the tactile simulation by EIP, we further propose a tactile-visual perception network that enables information fusion between tactile data and visual images. The perception network is based on a global-to-local fusion mechanism where multi-scale tactile features are aggregated to the corresponding local region of the visual modality with the guidance of tactile positions and directions. The fusion method exhibits superiority regarding the 3D geometric reconstruction task. Our code for EIP is available at \url{https://github.com/yikaiw/EIP}.
\end{abstract}



\maketitle

\section{Introduction}
\label{intro}
Tactile sensing is one of the most compelling perception pathways for nowadays robotic manipulation, as it is able to capture the physical patterns including shape, texture, and physical dynamics that are not easy to perceive via other modalities, \emph{e.g.} vision. In recent years, data-driven machine learning approaches have exploited tactile data and exhibited  success in a variety of robotic tasks, such as object recognition~\cite{liu2017recent,huang2016sparse}, grasp stability detection~\cite{kwiatkowski2017grasp,zapata2019tactile}, and manipulation~\cite{fang2018dual,tian2019manipulation} to name some. That being said, the learning-based methods---particularly those involving deep learning---are usually data-hungry and require large datasets for training. Collecting a large real tactile dataset is not easy, since it demands continuous robot control which is time-consuming or even risky considering the hardware wear and tear. Another concern with the real tactile collection is that the data acquired by the sensors of different shapes/materials or under different control policies could be heterogeneously distributed, posing a challenge to fairly assess the effectiveness of different learning methods trained on different tactile datasets.  

\begin{figure}[t]
\centering
\vskip 0.1 in
\hskip -0.03 in
\includegraphics[scale=0.4]{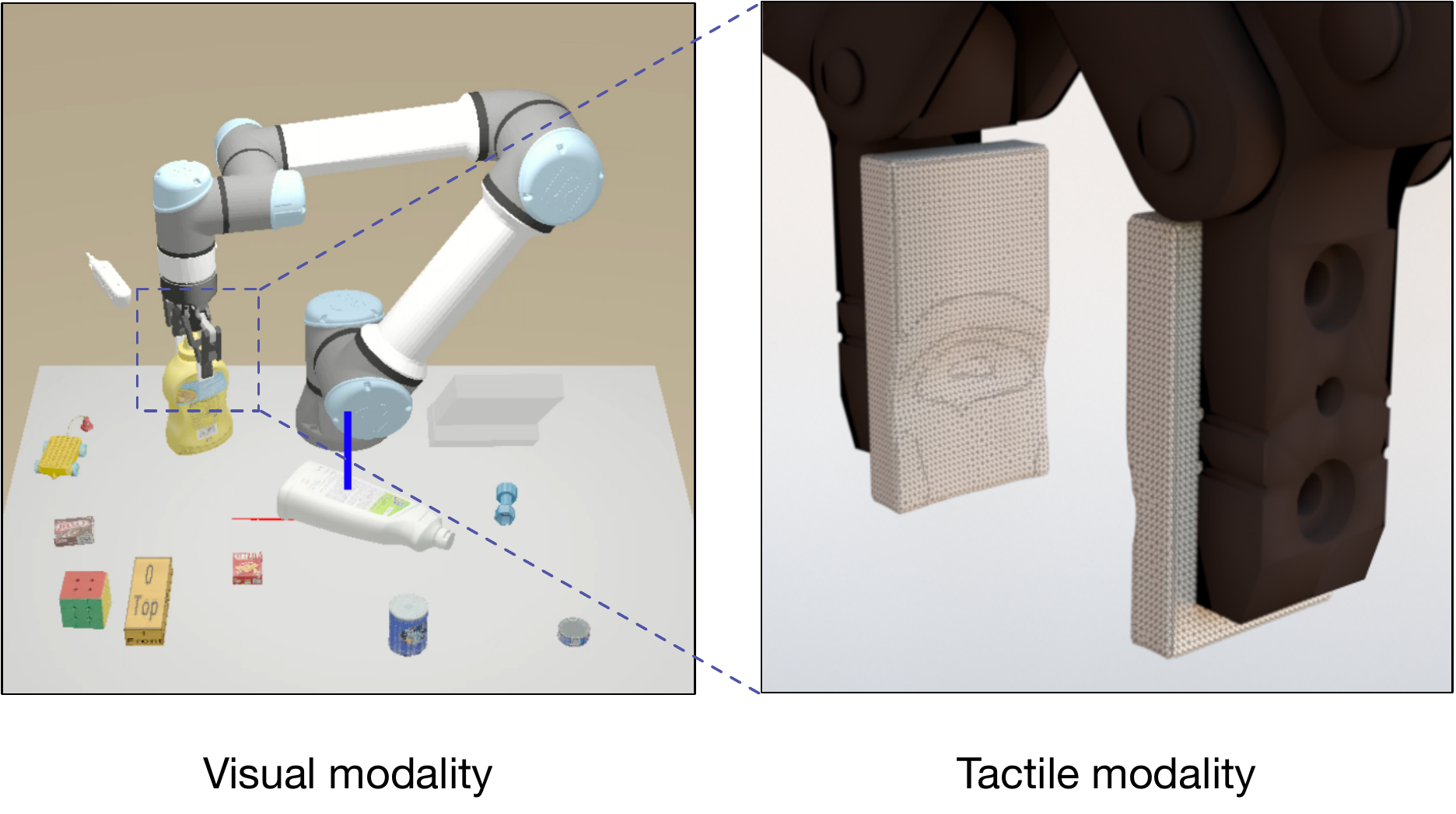}
\vskip -0.05 in
\caption{\textmd{Illustration of the simulated tactile pattern when a robot is grasping a mustard bottle. The framework enables researchers to deal with tactile-visual perception in the simulated environment.}}
\label{fig:example}
 \vskip -0.1 in
\end{figure}

\renewcommand*{\thefootnote}{\fnsymbol{footnote}}
\footnotetext[0]{\small \textsuperscript{\Letter} Corresponding author: Fuchun Sun.}
\renewcommand*{\thefootnote}{\arabic{footnote}}

The simulation of tactile sensing can potentially help overcome these real-world limitations. Yet, establishing a promising tactile simulator is challenging since the tactile sensor needs to be geometrically and physically modeled. In addition, we should be capable of characterizing the physical interaction during the contact process between the sensor and the object which makes simulating tactile data more difficult than other modalities to some extent, such as vision that is solely geometrically aware.

Existing trails that consider simulating tactile interactions with manipulated objects~\cite{moisio2013model,kappassov2020simulation,sferrazza2020learning,ding2020sim} usually model the tactile sensor as a combination of rigid bodies, and the collision between two objects is described by rigid multi-body kinematics provided by certain off-the-shelf physical engines (such as ODE in~\cite{kappassov2020simulation}). Despite its validity in some cases, considering the tactile sensor as a rigid multi-body will overlook the fact that common tactile sensors are usually elastic but not rigid. For example, the sensors invented by~\cite{yuan2017gelsight} leverages elastic materials to record the deformation to output tactile sensing. Moreover, in current methods, the segmentation of tactile sensor into rigid bodies is usually coarse and the interaction between rigid bodies is unable to capture the high-resolution sensor-object contact.

In this paper, we propose a novel methodology for tactile simulation, dubbed as Elastic Interaction of Particles (EIP). EIP first models the tactile sensor as a group of coordinated particles of certain mass and size. By assuming the sensor to be made of elastic materials, the elastic property is applied to constraint the movement of particles~\cite{stomakhin2012energetically}. During the interaction between the sensor and the object, the deformation of particles is recorded as tactile data. An example of the simulated tactile perception is illustrated in Figure~\ref{fig:example}.

With fine-grained  tactile patterns available, in this paper, we further propose a tactile-visual perception method that densely fuses features of both modalities, which exhibits great advantages on the 3D geometric reconstruction task. Multimodal learning can exhibit remarkable benefits against the unimodal paradigm if the patterns of different modalities are aligned and aggregated desirably~\cite{ramachandram2017deep,journals/ijcv/ValadaMB20,wang2020asymfusion}. However, tactile signals and visual images are not naturally aligned, since these two modalities are collected separately in different viewpoints. The discrepancy of viewpoints leads to geometrically unaligned tactile-visual feature maps.
To this end, we design a global-to-local fusion network to integrate the learned  tactile features into the visual counterparts. Since our tactile simulation is able to provide the position and direction for each touch,  each feature map of the tactile modality is  pooled into a global embedding and  is then located to a local visual region for feature aggregation. The tactile-visual fusion is conducted densely in the architecture to capture multi-scale resolutions. Finally, inspired by the scheme in Pixel2Mesh~\cite{wang2018pixel2mesh}, the aggregated features of both modalities are further sent to a GNN-based network for vertices deformation.

To sum up, our contributions are two-fold:
\begin{itemize}
    \item We propose EIP, a novel tactile simulating framework that is capable of modeling the elastic property of the tactile sensor and the fine-grained physical interaction between the sensor and the object. In contrast to existing methods that usually exploit the off-the-shelf physics engine for interaction simulation, the implementation of our method is formulated from scratch, which makes our framework more self-contained and easier to be plugged into downstream robotic applications. 
    \item We propose a global-to-local fusion method to densely aggregates features of tactile and visual modalities. The designed per-pixel fusion method and the feature  aggregation  process alleviate the misalignment issue of tactile-visual features. Experimental results on 3D geometric reconstruction support the effectiveness of the proposed scheme. We combine EIP with a robotic grasping environment to acquire real-time tactile signals of the manipulated objects, which verifies its potential to downstream robotic tasks. 
\end{itemize}

\section{Related Work}
\textbf{Tactile simulaton.}
The vision-based tactile sensors have become prominent due to their superior performance on robotic perception and manipulation. Data-driven approaches to tactile sensing are commonly used to overcome the complexity of accurately modeling contact with soft materials. However, their widespread adoption is
impaired by concerns about data efficiency and the capability to generalize when applied to various tasks.
Hence simulation approaches of vision-based tactile sensing are developed recently. Regarding the exploration of tactile simulation, early work \cite{journals/trob/ZhangC00} that directly adopts the elastic theory for the mesh interaction resorts to high computation costs. 
\cite{moisio2013model} represents the tactile sensor as a rigid body and calculates the interaction force on each triangle mesh.  
\cite{habib2014skinsim} models the tactile sensor as rigid elements and simulates their displacement by adding a virtual spring, with help of the commonly used Gazebo simulator. Modeling the tactile sensor as one or a combination of independent rigid bodies makes these methods difficult to obtain high-resolution tactile patterns, and these methods also overlook the fact that tactile sensors are mostly elastic materials. 
\cite{ding2020sim} implements the soft body simulation  based on the Unity physics engine and trains a neural network to predict the contact information including positions and angles.  \cite{gomes2019gelsight} introduces an approach for simulating a GelSight tactile sensor in the Gazebo simulator by directly modeling the contact surface, which yet neglects  the elastic material of the tactile sensor. Based on the finite-element analysis, \cite{sferrazza2020learning} provides a simulation strategy to generate an entire supervised learning dataset for a vision-based tactile sensor, intending to estimate the full contact force distribution from real-world tactile images.

\textbf{Tactile-visual perception.} This paper mainly discusses the application of tactile-visual perception for the 3D reconstruction task. Several previous works combine  vision and touch for shape reconstruction which rely on the given point cloud and depth data ~\cite{DBLP:conf/iros/BjorkmanBHK13,DBLP:conf/icra/Watkins-VallsVA19,DBLP:journals/ijrr/IlonenBK14,DBLP:journals/ras/GandlerEBSB20}. Under such circumstances, shape reconstruction is less challenging since the real depth data or sparse point cloud at hand directly provides the global 3D information. When global depth is not available, \cite{DBLP:journals/corr/abs-1808-03247} proposes to first estimate the depth and normal direction based on the vision and then predicts 3D structure with shape priors. Real-world tactile signals subsequently refine the predicted structure. Instead of predicting the global depth from the visual perception, \cite{DBLP:conf/nips/SmithCRGMMD20} focuses on generating the local depth and point cloud from tactile signals, which provides local guidance for the chart deformation when a mass of tactile signals are obtained. \cite{sundaram2019learning,edmonds2019tale} discuss the tactile-visual perception based on real robotic hands from both
functional and mechanistic perspectives. Different from existing methods, our tactile-visual perception framework is self-contained and end-to-end trainable without leveraging the global/local depth or point cloud information. We focus on how to combine geometrically unaligned feature maps for improving  fusion performance.

\begin{figure*}[t]
\centering
\includegraphics[scale=0.57]{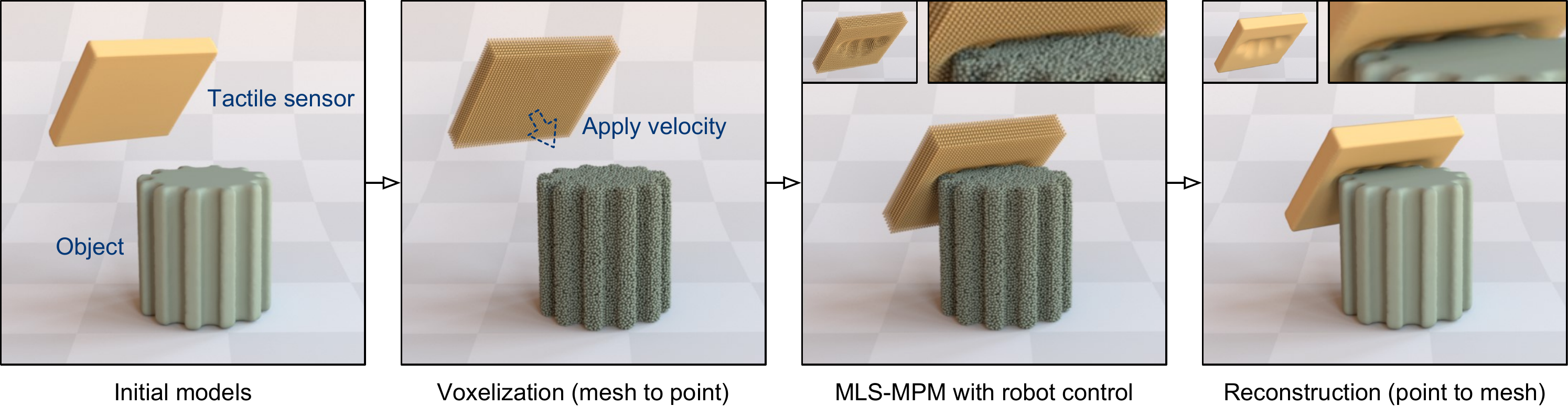}
\caption{\textmd{A brief framework of our tactile simulation process. The initial 3D meshes are first converted to particles by voxelization. Particles of the tactile sensor and the manipulated object are interacted by MLS-MPM, with additional control at the robot side. Meshes can be optionally  reconstructed from the points for rendering purposes.}}
\label{fig:channel}
\end{figure*}

\section{Physically Tactile Simulation}

In this section, we first introduce how to simulate the physical interaction between the tactile sensor and the targeted object. 
We then evaluate the effectiveness of our simulated tactile sensor by 3D geometric reconstruction with tactile-visual fusion. 

\subsection{Overall Framework}

The basic idea of our method is to assume both the tactile sensor and the manipulated object to be solid and model them in the form of particles. Differently, the sensor and the object are considered elastic and rigid respectively. In general, the tactile simulation process depicted in Figure~\ref{fig:channel} consists of 3 steps: voxelization from meshes to particles, interaction simulation, and reconstruction from particles to meshes. We detail each step below.

\paragraph{Voxelization}\vskip-0.3em
We first obtain the triangle meshes from the simulation environment that describe the geometric model of the  object/sensor. The inside of each model is filled with dense voxel grids by voxel carving. Briefly, we first calculate the depth maps and then employ these depth maps to carve a dense voxel grid. We refer readers to \cite{Zhou2018} for more details. The center of each voxel grid is denoted as a particle, which is the fundamental unit for the following physical interaction simulation.

\paragraph{Interaction Simulation}\vskip-0.3em
After voxelization, we apply a certain velocity to the tactile sensor until it touches the object to a certain extent. The interaction process can be represented by the deformation of particles in the tactile sensor. Thus we simulate the deformation process based on Material-Point-Method (MPM)~\cite{stomakhin2013material} and its modification MLS-MPM~\cite{hu2018moving} considering both efficacy and efficiency, where we simultaneously apply the specific movement of the tactile sensor under robot control. Details of this step are provided in \textsection~\ref{sec:interaction}.  

\paragraph{Reconstruction}
The final step is to reconstruct the meshes based on the positions of particles, which can be accomplished by using the method proposed by~\cite{kazhdan2013screened}. Note that this step is not necessary unless we want to render the interaction at each time step.

\subsection{Interaction Simulation}
\label{sec:interaction}
This subsection presents details of how we simulate the interaction process. 
In practice, the tactile sensor is basically made of elastic materials, and our main focus is on the change of its shape, or called deformation. We apply the elastic theory to constrain the deformation of the particles in the tactile sensor during its interaction with the manipulated object, and the deformation at each time step will be recorded as the tactile sensing data.

Suppose that the sensor is composed of $m$ particles. The coordinate of the $p$-th particle is represented by $\vx_p\in\R^{d}$, where $d=3$ throughout our paper. We define the deformation map as $\bm{\Phi}: \R^{d}\rightarrow \R^{d}$. The Jacobian of $\bm{\Phi}$ with respect to the $p$-th particle, denoted as $\mF_p\in\R^{d\times d}$ (\emph{a.k.a} deformation gradient), is calculated by
\begin{eqnarray}
\label{eq:deformation-gradient}
\mF_p = \frac{\partial \bm{\Phi}}{\partial \vx}(\vx_p).
\end{eqnarray}
When the particle deforms, its volume may also change. The volume ratio by the deformation, denoted as $J_p$, is the determinant of $\mF_p$, \emph{i.e.}, 
\begin{eqnarray}
J_p=\mathrm{det}(\mF_p).
\end{eqnarray}

To describe the stress-strain relationship for elastic materials, we adopt a strain energy density function $\bm{\Psi}$, a kind of potential function that constrains the deformation $\mF_p$. We follow a widely used method called Fixed Corotated~\cite{stomakhin2012energetically}, which computes $\bm{\Psi}$ by
\begin{eqnarray}
\label{eq:phi}
    \bm{\Psi}(\mF_p) =\mu\sum_{i=1}^d(\sigma_{i,p}-1)^2+\frac{\lambda}{2}(J_p-1)^2,
\end{eqnarray}
where $\mu=\frac{E}{2(1+v)}$ and $\lambda=\frac{Ev}{(1+v)(1-2v)}$ are Lamé's 1st and 2nd parameters, respectively; $E$ and $v$ are Young's modulus and Poisson ratio of the elastic material, respectively; $\sigma_{i,p}$ is the $i$-th singular value of $\mF_p$. The derivative of $\bm{\Psi}$ (\emph{a.k.a} the first Piola-Kirchhoff stress) will be utilized to adjust the deformation process, derived by
\begin{eqnarray}
\label{eq:PK}
    \mP_p=\frac{\partial \bm{\Psi}}{\partial \mF}(\mF_p)=2\mu(\mF_p-\mR_p)+\lambda(J_p-1)J_p\mF_p^{-\mathsf{T}},
\end{eqnarray}
where $\mR_p$ is obtained via the polar decomposition~\cite{higham1986computing}: $\mF_p=\mR_p\mS_p$.

In the following context, we will characterize how each particle deforms, that is, how its position $\vx_p$ changes during the interaction phase. For better readability, we distinguish the position $\vx_p$ and the quantities in Eq.~(\ref{eq:deformation-gradient}-\ref{eq:PK}) at each different time step by adding a temporal superscript, \emph{e.g.} denoting the velocity at time step $n$ as $\vx_p^{(n)}$. We leverage MLS-MPM~\cite{hu2018moving} to update $\vx_p^{(n)}$, which divides the whole space into grids of a certain size. For each particle, its velocity is updated as the accumulated velocity of all particles within the same grid, which to some extent can emulate the physical interaction between particles. 
Specifically, we iterate the following steps for the update of $\vx_p^{(n)}$. The flowchart is sketched in Algorithm~\ref{alg:tactile_simulation}.

\paragraph{Momentum Scattering}
For each grid, we collect the mass and the momentum from the particles inside and those within its neighbors. The mass of the $i$-th grid is collected by 
\begin{equation}
\label{eq:mass}
    m'_i = \sum_{j\in\sG_i}\sum_{p\in\sP_j} w_{jp}m_p,
\end{equation}
where $\sG_i$ denotes the $3\times3\times3$ neighbor grids surrounding grid $i$; here, only considering the effects of particles in $\sG_i$ and omitting other distant particles are due to the computational efficiency; $\sP_j$ collects the indices of the particles located in grid $j$; $m_p$ denotes the mass of the $p$-th particle; $w_{jp}$ computes the B-Spline kernel negatively related to the distance between the $j$-th grid and the $p$-th particle. 

The momentum of the $i$-th grid is derived as
\begin{align}
\label{eq:momentum}
(m'\vv')_i &=\sum_{j\in\sG_i}\sum_{p\in\sP_j} w_{jp} \left(m_p\vv_p^{(n)}+\mC_p^{(n)}(\vx'_j-\vx_p^{(n)})\right)\notag\\
&-\sum_{j\in\sG_i}\sum_{p\in\sP_j} w_{jp}\gamma\mP_p^{(n)}(\mF_p^{(n)})^{\mathsf{T}}(\vx'_j-\vx_{p}^{(n)}),
\end{align}
where $\vx'_j$ denotes the position of grid $j$; $v_p^{(n)}$ is the velocity of particle $p$;  $\mC_p^{(n)}\in\R^{d\times d}$, adopted as an approximated parameter in~\cite{hu2018moving}, is associated to the particle $p$ whose update will be specified later; $\gamma=\frac{4\Delta t}{\Delta x'^2} V_p^0$ is a fixed coefficient where $\Delta t$ is the time interval; $\Delta x'$ is the spatial interval between grids;  $V_p^0$ is the initial particle volume. 

\paragraph{Velocity Alignment}
The velocity on the $i$-th grid can be obtained given the grid momentum and the grid mass by normalization, \emph{i.e.},
\begin{eqnarray}
\label{eq:grid_velocity}
    \vv'_i=\frac{(m'\vv')_i}{m'_i}.
\end{eqnarray}
Note that the grid velocity is only for later parameter updates, and the position of the grid will not change in the simulation. 

\begin{figure}[t]
\centering
\includegraphics[scale=0.42]{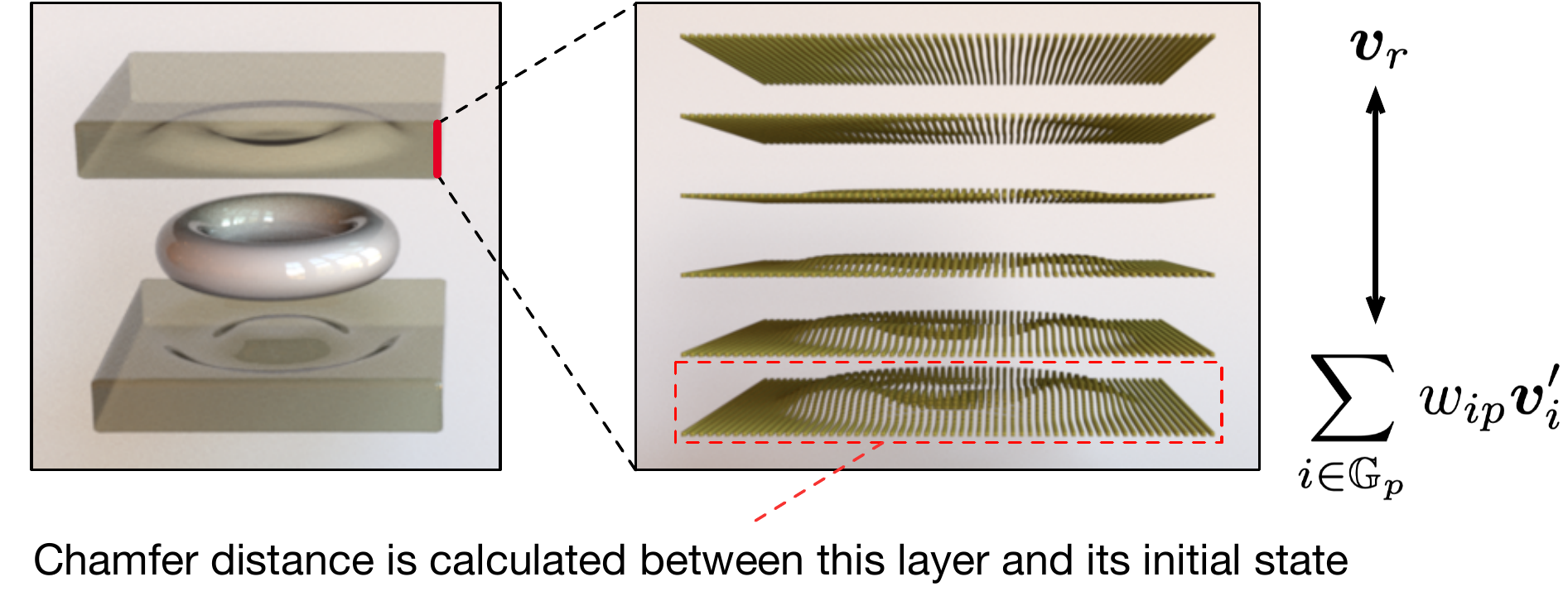}
\vskip -0.05 in
\caption{\textmd{Deformation of pressing a torus mesh.  To describe the transition from the velocity of the robot hand to the particle interacted velocity, we depict the deformation of each layer. The layer that directly contacts with the manipulated object has the largest extent of deformation, and we also use this layer to calculate the chamfer distance for terminal checking. Note that we increase the distance of tactile sensors for better visualization.}}
\label{pic:chamfer}
\vskip -0.03 in
\end{figure}

\begin{figure}[t]
\centering
\includegraphics[scale=0.42]{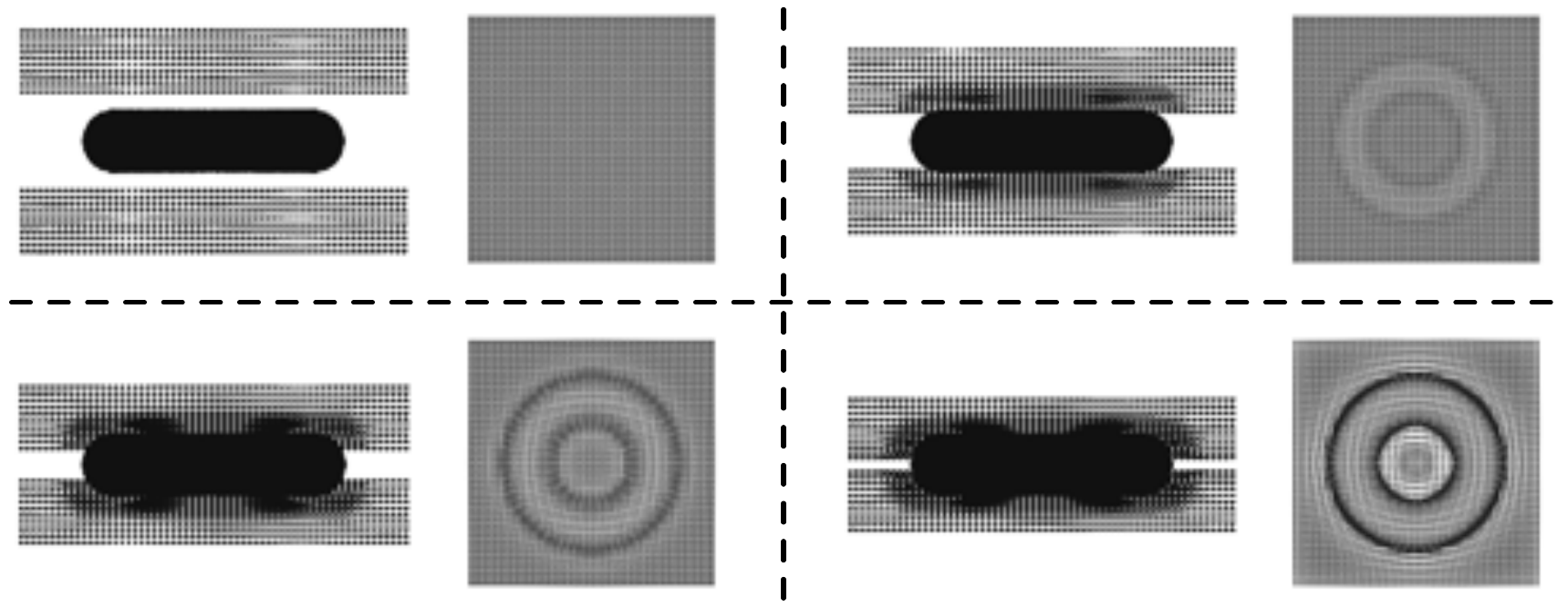}
\vskip -0.05 in
\caption{\textmd{The change of tactile deformation during pressing a torus mesh,  depicted using raw data.}}
\label{pic:deformation}
\vskip -0.03 in
\end{figure}

\begin{figure*}[t]
\centering
\vskip 0.1 in
\includegraphics[scale=0.46]{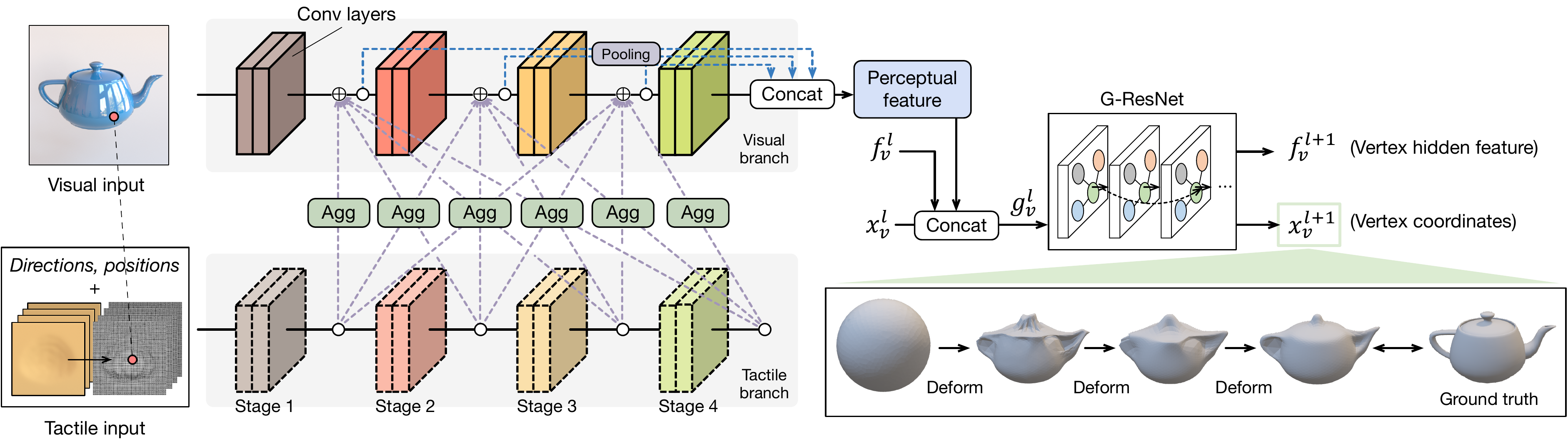}
\vskip 0.03 in
\caption{\textmd{Training process of the proposed tactile-visual fusion method. Feature maps of the tactile branch are densely merged to the visual branch with Aggregation (Agg) blocks. Here, Agg transforms each tactile feature map into a $1\times1$ embedding, which is further added to $1\times1$ embeddings at the corresponding pixels in the visual feature map. More details are introduced in Figure~\ref{pic:agg}. Note that there is one Agg on each connection, and only one is depicted at intersections for visualization purposes.  In the visual branch, multi-layer features are collected as the perceptual feature, which is  then sent to the GNN (specifically, G-ResNet). Here, the average pooling technique is adopted for downsampling so that features from different stages can be concatenated. $\bigoplus$ indicates channel-wise addition of $1\times1$ embeddings. The learning process regarding GNN follows the method in Pixel2Mesh~\cite{wang2018pixel2mesh}. We provide an illustration of a teapot deforming from the initial sphere to the final prediction. }}
\label{pic:projection}
\end{figure*}

\begin{figure}[t]
\centering
\vskip 0.02 in
\hskip -0.02 in
\includegraphics[scale=0.5]{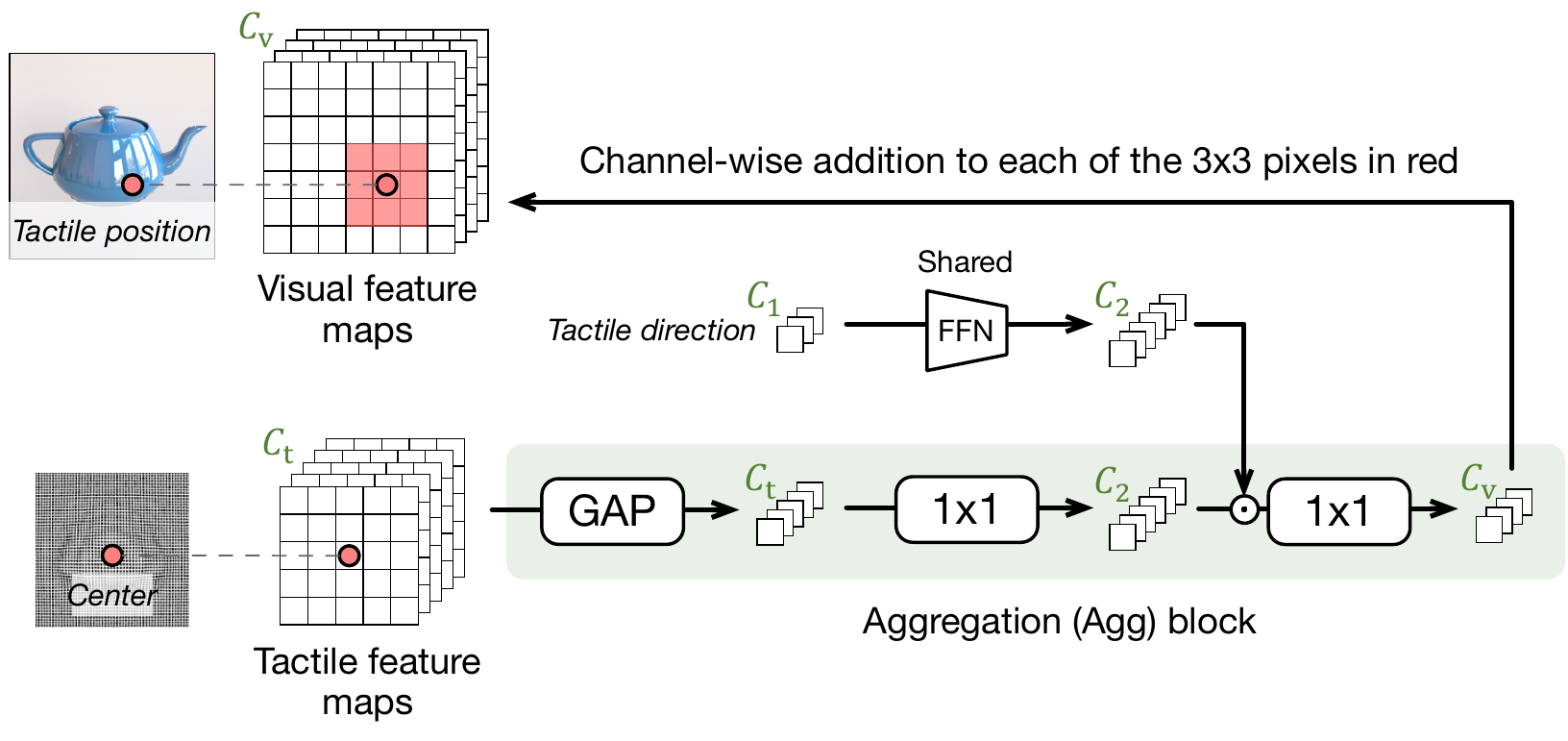}
\caption{\textmd{Illustration of the per-pixel fusion method and the feature transformation process with an Aggregation (Agg) block. GAP, $1\times1$ and FFN represent global average pooling, the $1\times1$ convolutional layer, and the feed-forward neural network, respectively. The channel number of each feature (map) is  annotated in its upper left. $\bigodot$ indicates channel-wise dot production. Each aggregated tactile embedding is added to the $3\times3$ visual pixels which are colored in red, which is for amplifying the multimodal fusion effectiveness.}}
\label{pic:agg}
\vskip -0.1in
\end{figure}

\paragraph{Robot Hand Movement}
Apart from the displacement caused by physical interaction, the sensor will also change its position due to the movement by the robot hand where the sensor is equipped. This additional velocity is a local parameter depending on the position of a particle, and we use a weight $\alpha_p$ to reflect it during the scattering process while keeping its sufficient physical interaction at the contact side. The velocity is thus updated as 
\begin{align}
\label{eq:particle_v}
\vv_p^{(n+1)}&=\alpha_p\sum_{i\in\sG_p} w_{ip}\vv'_i+(1-\alpha_p)\vv_r,
\end{align}
where $\sG_p$ is the set of $3\times3\times3$ grids that the particle $p$ adheres to; $\vv_r$ is the velocity of the robot hand;  $\alpha_p$ is proportional to the distance between the particle and the robot hand as illustrated in Figure~\ref{pic:chamfer}, and the deformation process is provided in Figure~\ref{pic:deformation}.

\paragraph{Parameters Gathering}
With the updated velocity, we renew the values of the velocity gradient, position vector, and deformation gradient by
\begin{align}
\label{eq:particle_c}
\mC_p^{(n+1)}&=\frac{4}{\Delta x^2}\sum_{i\in\sG_p} w_{ip}\vv_p^{(n+1)}(\vx_i-\vx_p^{(n)}),\\
\label{eq:particle_x}
\vx_p^{(n+1)}&=\vx_p^{(n)} + \Delta t \vv_p^{(n+1)},\\
\label{eq:particle_f}
\mF_p^{(n+1)}&=(\mI + \Delta t \mC_p^{(n+1)})\mF_p^{(n)}.
\end{align}

\begin{algorithm}[t]
	\small
	\caption{\small{Elastic Interaction of Particles (EIP)}}
	\begin{algorithmic}[1]
		\Require 3D meshes of the manipulated object and the tactile sensor; values of Lamé's parameters including $E$ and $v$; robot hand velocity $\vv_{r}$.
		\Ensure Tactile interaction between the object and the sensor.
		\State Convert meshes to particles using voxelization.
		\State Initialize the values of $\vx_p^{(0)}$, $\mF_p^{(0)}$, $\mC_p^{(0)}$, and $\vv_p^{(0)}$. 
		\State Dividing the whole space into grids.
		\While{not terminal}
		\For{each grid $i$}
		\State Scatter the mass and momentum of grid $i$ by Eq.~(\ref{eq:mass}-\ref{eq:momentum}).
		\State Update the grid velocity by Eq.~(\ref{eq:grid_velocity}).
		\EndFor
		\For{each particle $p$}
		\State Gather the velocity $\vv_p^{(n)}$ by Eq.~(\ref{eq:particle_v}).
		\State Update parameters $\vx_p^{(n)}$, $\mF_p^{(n)}$, and $\mC_p^{(n)}$ by Eq.~(\ref{eq:particle_c}-\ref{eq:particle_f}).
		\EndFor
		\State  Terminal check of the robot control by Eq.~(\ref{eq:chamfer}).
		\EndWhile

	\end{algorithmic}
	\label{alg:tactile_simulation}

\end{algorithm}

\paragraph{Terminal Checking}
For safety and keeping consistent with the practical usage, we will terminate the robot hand movement once the deformation of the sensor is out of a certain scope. For this purpose, we use the chamfer distance to measure the distance between the deformed state and the original state of the particles in the contact surface, also illustrated in Figure~\ref{pic:chamfer}. In form, we compute
\begin{equation}
\label{eq:chamfer}
   l=\sum_{p\in\sS} \min_{q\in\sS}\|\hat{\vx}_p^{(n+1)}-\hat{\vx}_q^{(0)}\|_2^2+\sum_{q\in\sS} \min_{p\in\sS}\|\hat{\vx}_p^{(n+1)}-\hat{\vx}_q^{(0)}\|_2^2,
\end{equation}
where $\sS$ denotes the contact surface between the sensor and the object; $\hat{\vx}=\vx-\bar{\vx}$ for removing the effect of translation brought by $\vv_r$, and $\bar{\vx}$ denotes the center point of $\vx$.

\begin{figure*}[t]
\centering
\hskip 0.09 in
\includegraphics[scale=0.14]{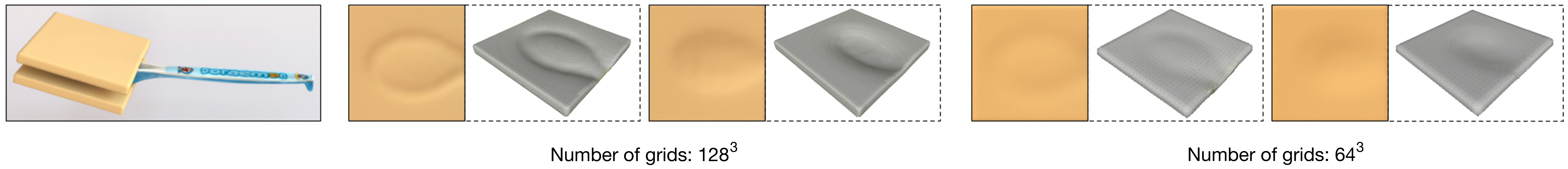}
\vskip -0.05 in
\caption{\textmd{Comparison of tactile patterns when pressing a spoon, with different grid number settings. }}
\label{pic:spoon}
\end{figure*}

\begin{figure}[t]
\centering
\includegraphics[scale=0.37]{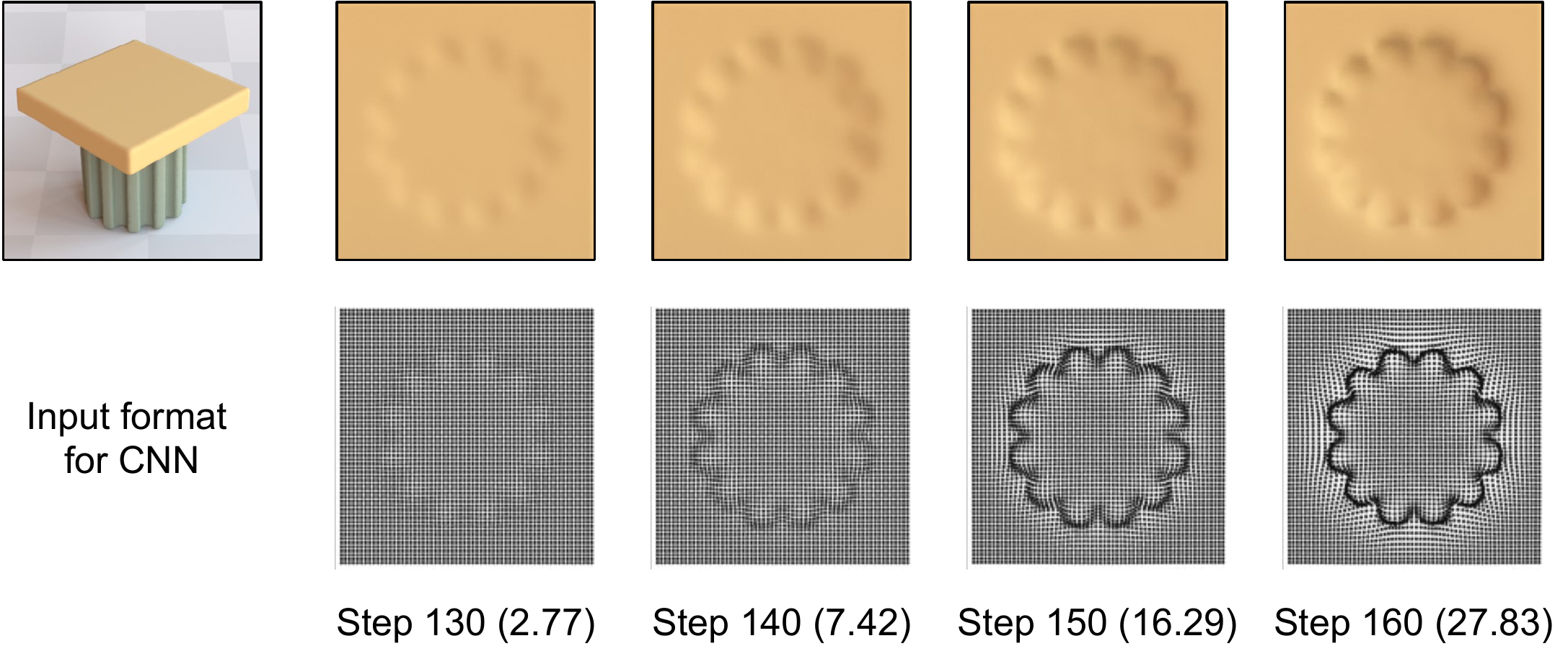}
 \vskip -0.05 in
\caption{\textmd{Comparison of tactile patterns at different time steps. The extent of deformation increases with the time step increases, followed by the increase of the chamfer distance $l\;(10^{-5})$ as shown in the brackets. We also depict the input format for CNN.}}
\label{pic:extent}

\end{figure}

\begin{figure}[t]
\centering
\includegraphics[scale=0.37]{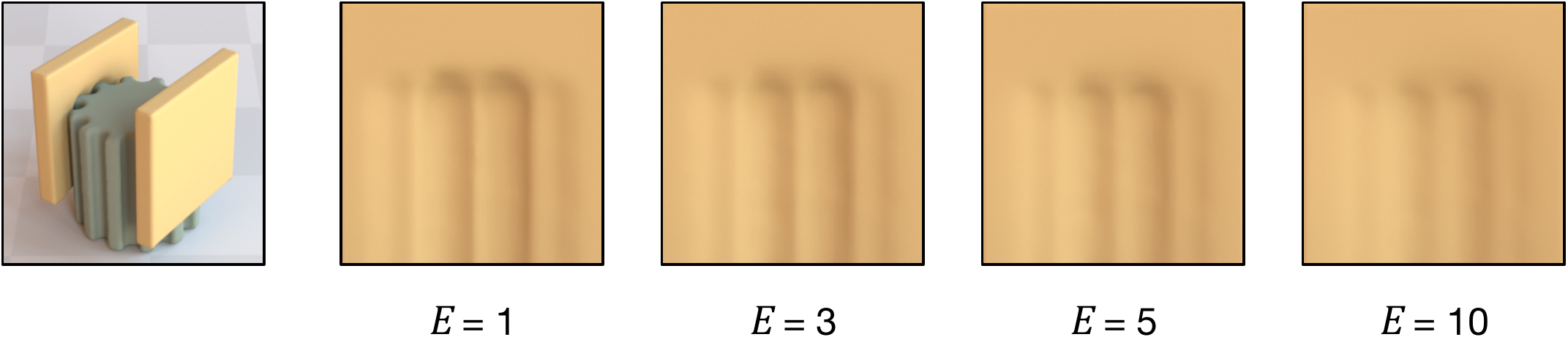}
\vskip -0.05 in
\vskip -0.05 in
\caption{\textmd{Comparison of tactile patterns with different elastic coefficients (Young’s modulus $E$), at the same time step.}}
\label{pic:young}
\vspace{-0.5em}
\end{figure}

\subsection{Tactile-Visual Multimodal Fusion}
\label{sec:application}

CNN-based architectures have shown the superiority in homogeneous multimodal fusion, \emph{e.g.}, RGB-depth fusion~\cite{wang2020cen}, where feature maps of different modalities are naturally aligned. But here, the feature maps of vision and touch are unaligned due to the discrepancy of viewpoint and tactile direction. Existing methods mostly leverage estimated global or local depth for multi-modal alignment. In this section, we propose an end-to-end tactile-visual perceptual framework that directly combines multi-modal features without explicitly predicting depth or sparse point cloud.

Specifically, we focus on the 3D geometric reconstruction of the manipulated objects, where we utilize the simulated tactile signals as complementary information to single-view visual images. During the prediction process, the network input is composed of an image and a set of tactile data obtained from different grasps based on different directions. 

The structure of our tactile-visual perception framework is illustrated in Figure~\ref{pic:projection}. We adopt a visual branch and a tactile branch to process visual and tactile inputs, respectively. We choose ResNet~\cite{he2016deep} as the backbone network for both branches, and thus there are four stages depicted. Multi-scale tactile features are extracted and densely connected to  each network stage in the visual branch for multimodal feature fusion. We then concatenate multi-stage feature maps as a perceptual feature, which is finally sent to a Graph Neural Network (GNN) to predict the vertices deformation of the 3D mesh following the scheme in Pixel2Mesh~\cite{wang2018pixel2mesh}.

The main contributions of our  method include a designed Aggregation (Agg) block and the global-to-local feature fusion method to alleviate the misalignment issue.  As illustrated in Figure~\ref{pic:agg}, Agg transforms tactile features before integrating them into the visual branch. The transformation consists of a Global Average Pooling (GAP), which results in a global embedding, and two $1\times1$ Conv layers for feature extraction. Given that the touch direction plays an important role in tactile representation, we additionally involve the direction information into our tactile processing, by using a Feed-Forward Network (FFN) (that is shared in different Agg blocks) to produce the direction embedding which is integrated with the tactile global embedding via dot product. Note that in Agg, each $1\times1$ Conv layer is followed by a ReLU activation and a batch normalization layer.
As described in Sec.~\ref{intro}, the tactile-visual feature maps are not naturally aligned due to the discrepancy of viewpoint and tactile direction. In our tactile generation, we are able to record the position (\emph{i.e.} the coordinate center of tactile pixels). With the help of the tactile position, we then locate the corresponding local region (the $3\times3$ region colored in red in Figure~\ref{pic:agg}) in each visual feature map and then perform addition between the global tactile embedding by Agg and every pixel within the local visual region. It is known that the different-stage layer in neural networks characterizes different-scale patterns. Hence, we conduct the above tactile-visual Agg between every two stages across the tactile and visual branches, to enable multi-scale and delicate fusion. 

Inspired by~\cite{wang2018pixel2mesh}, the perceptual feature is sent to a GNN to guide the deformation process of GNN vertices, which are also the vertices of the 3D geometric model to be reconstructed. The geometric model is initialized to an ellipsoid mesh and the deformation is realized by updating 3-dimensional vertices embeddings (coordinates) of GNN. Each vertex on the mesh would be projected to the closest visual feature~\cite{wang2018pixel2mesh}. The deformation update process is illustrated in the right part of Figure~\ref{pic:projection}, and can be formulated as below,
\begin{align}
\vg_v^l&=\mathrm{concat}\left(\vf_v^l,\vx_v,\text{proj}(\vf_{p})\right),\\
\vf_{v}^{l+1}&=\mW_{f0}\vg_v^l+\sum_{v'\in\mathcal{N}(v)}\mW_{f1}\vg_{v'}^l,\\
\vx_{v}^{l+1}&=\mW_{x0}g_v^l+\sum_{v'\in\mathcal{N}(v)}\mW_{x1}\vg_{v'}^l,
\end{align}
where $\vf_{p}$ denotes the perceptual feature learned by tactile-visual fusion, which has been introduced above (annotated in a blue box in Figure~\ref{pic:projection}); $v$ denotes the index of the vertex on the mesh; $\mathcal{N}(v)$ denotes the neighbors of $v$, which is available as the mesh is initially a sphere; $\vf_v^l,\vx_v^l$ are the feature representation and the learned coordinate of vertex $v$ for the $l$-th GNN layer, respectively; $\vg_v^l$ is the hidden feature given by the concatenation of $\vf_v^l,\vx_v$ and a multimodal feature projection on $\vf_{p}$, denoted as $\text{proj}(\vf_{p})$; $\mW_{f0},\mW_{f1},\mW_{x0},\mW_{x1}$ are learnable weights.

\vspace{0.5em}
\section{Experiment}
We implement Algorithm~\ref{alg:tactile_simulation} based on Taichi~\cite{hu2019taichi}, and  adopt Mitsuba~\cite{nimier2019mitsuba} for rendering 3D models, \emph{e.g.} in Figure~\ref{fig:channel} and Figure~\ref{pic:spoon}.

\subsection{Effects of Coefficient Settings in EIP}
In Figure \ref{pic:spoon}, we provide the tactile patterns when pressing a spoon, and compare the patterns under different grid numbers (described in \textsection~\ref{sec:interaction}). We observe that the larger the number of grids is the more fine-grained simulation we will attain. We set the grid number as $128^3$ considering the trade-off between efficacy and efficiency. To illustrate how the deformation behaves during the  contact, in Figure \ref{pic:extent}, we keep pressing the tactile sensor on a gear object and record results at different time steps. For each time step, we also provide its corresponding format for CNN input, and the Chamfer distance $l$ as described in \textsection~\ref{sec:interaction}; The deformation  becomes more remarkable as the contact proceeds.  Figure \ref{pic:young} contrasts the influence by Young's modulus $E$ under the same press displacement. It is shown that the tactile range gets smaller with the increase of  Young's modulus, which is consistent with the conclusion in elastic theory. In our simulation, we choose $E=3$ and $v=0.25$ in Eq.~\ref{eq:phi} by default.

\subsection{Tactile Dataset}
\begin{figure}[t]
\centering

\includegraphics[scale=0.265]{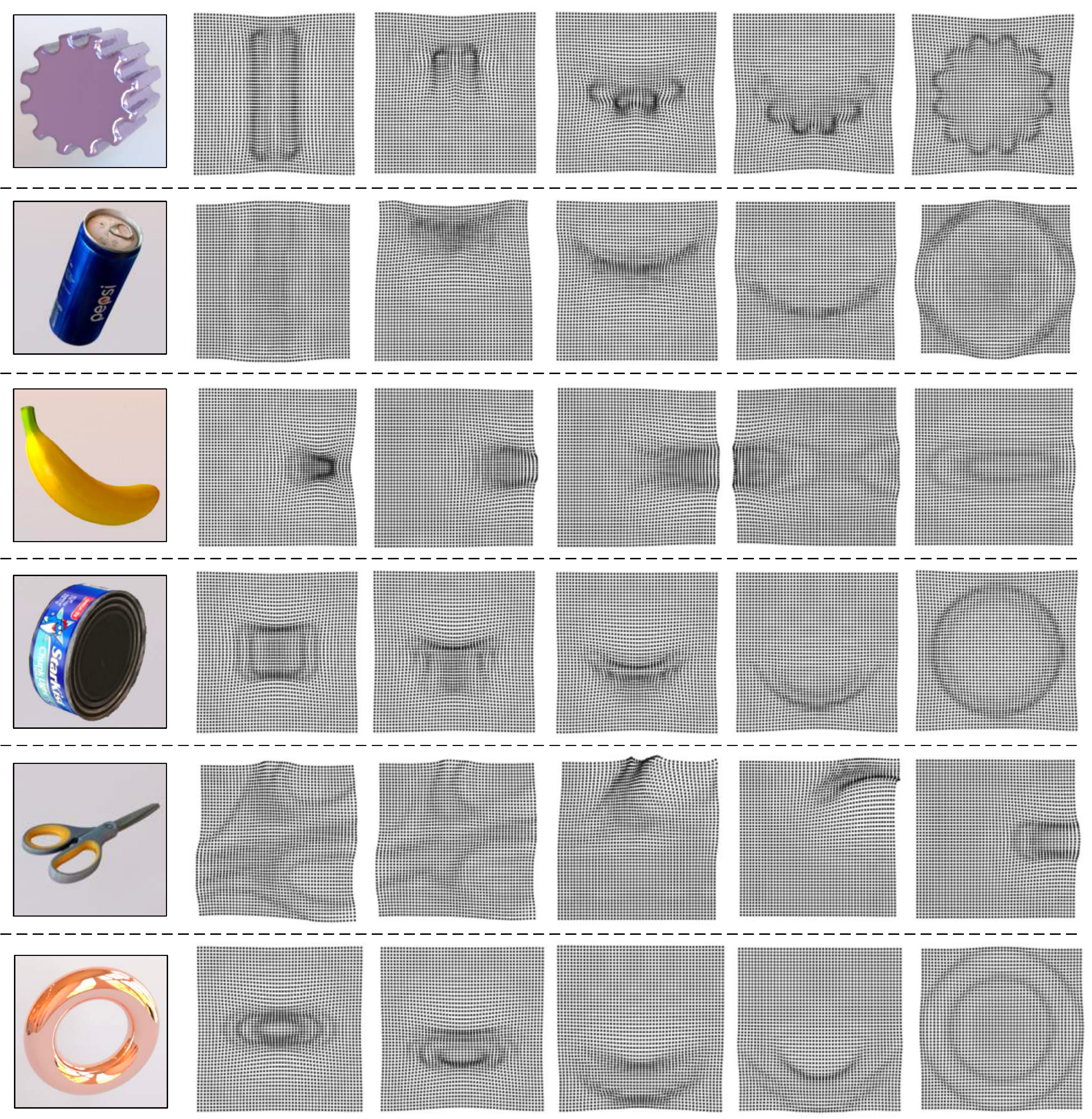}
\vskip -0.1 in
\caption{\textmd{Sample visualization of our tactile dataset generated by EIP simulation.}}
\label{pic:data_example}
\end{figure}

We build a tactile dataset containing 7,000 tactile images of 35 different object classes. These tactile images are collected through different contact policies including press directions and forces. The dataset summary and train/test splits are provided in Table~\ref{tabs:dataset}. Figure \ref{pic:data_example} illustrates an example subgroup of the tactile dataset.

Once we obtain the tactile data (namely, the deformation of particles of the sensor), we can apply these data for object recognition which partially evaluates the quality of the tactile data. We suppose the tactile deformation of the contact surface as $\mX^{(N)}\in\R^{H\times W\times d}$, where $H$ and $W$ denote the height and the width of the sensor, respectively, and $N$ denotes the final time step. Then, we train a neural network $\hat{y} = f(\mX^{(N)})$ to predict the object label. In practice, we prefer to try several attempts of the touch for more accurate recognition. All the deformation outcomes of different touching, denoted as $\{\mX_i^{(N)}\}_{i=1}^I$ will be concatenated along the channel direction, leading to $\mZ\in\R^{W\times H\times (Id)}$, as input of the network $f$. 

To predict the object category given tactile patterns, we train a ResNet-18~\cite{he2016deep} with ImageNet pretraining. We also conduct prediction experiments with more than one input tactile image. Specifically, during each training iteration, we randomly choose $N$ tactile images which are yielded by different press directions to the same object. As introduced in~\textsection~\ref{sec:application}, we treat the number of images as the number of the input channel. Table \ref{tabs:classification} 
summarizes the accuracies of the tactile perception with the number of input images (touches) from 1 to 10. It reads that increasing the number of  touches consistently improves the classification accuracy, and when the number is equal to 10, the accuracy becomes close to $93\%$, which implies the potential usage of our tactile simulation for real robotic perception.

\begin{table}[t]
\centering\caption{\textmd{Summary of the dataset structure and dataset split for downstream tasks. We augment the dataset with random mesh deformation to improve the object diversity which also reduces object intersections in training and testing. }}
\vskip -0.1 in
\tablestyle{2.5pt}{1}
\resizebox{80mm}{!}{
\begin{tabular}{c|c|ccc|ccc}
\toprule[0.25mm]
\multirow{2}*{Task}&\multirow{2}*{Resolution}&\multicolumn{3}{c|}{Training data}&\multicolumn{3}{c}{Testing data}\\

&&Class&Tacile & Visual&Class&Tacile  & Visual  \\

\midrule[0.15mm]
Classification&$224\times224$&35&5,500&0&35&1,500&0\\
3D reconstruction&$224\times224$&35&6,500&650&10&500&50\\
\bottomrule[0.25mm]
\end{tabular}}
\label{tabs:dataset}
\vskip -0.03 in
\end{table}

\begin{table}[t]
\centering\caption{\textmd{Classification accuracies (\%) of the tactile perception. Results of each setting are collected over three runs.}}
\vskip -0.1 in
\tablestyle{2pt}{1}
\resizebox{80mm}{!}{
\begin{tabular}{p{1cm}<{\centering}p{1cm}<{\centering}p{1cm}<{\centering}p{1cm}<{\centering}p{1cm}<{\centering}p{1.3cm}<{\centering}}
\toprule[0.25mm]
1-touch&2-touch&4-touch&6-touch&8-touch&10-touch  \\
\midrule[0.15mm]
36.7$\pm$2.3&47.4$\pm$1.6&59.7$\pm$1.1&81.5$\pm$0.7&90.2$\pm$0.6&92.9$\pm$0.3\\
\bottomrule[0.25mm]
\end{tabular}}
\label{tabs:classification}
\vskip -0.05 in
\end{table}

\begin{table}[t]
\centering\caption{\textmd{Results comparison of mesh reconstruction. Each result is the average over 10 different manipulated objects. Evaluation metric: chamfer distance ($\times10^{-3}$), lower values indicate better performance. }}
\vskip -0.1 in
\tablestyle{1pt}{1}
\resizebox{80mm}{!}{
\begin{tabular}{c|c|ccc|ccc}
\toprule[0.25mm]
\multirow{2}*{Model}&\multirow{2}*{Visual}&\multicolumn{3}{c|}{Tactile}&\multicolumn{3}{c}{Tactile \& Visual}\\
&&2-touch   &5-touch&10-touch&2-touch   &5-touch&10-touch  \\
\midrule[0.15mm]
Chart-based~\cite{DBLP:conf/nips/SmithCRGMMD20}&13.65&-&25.06&17.44&-&8.10&5.83\\
Ours&14.07&28.70&21.91&15.65&10.27&7.22&5.32\\
\bottomrule[0.25mm]
\end{tabular}}
\label{tabs:chamfer}
\end{table}

\begin{table*}[t]
\centering\caption{\textmd{Results comparison with SOTA method and ours on 10 manipulated objects. Ablation studies on our method have been also performed, and their detailed descriptions are provided in Sec~\ref{sec:exp_mesh}. Evaluation metric: chamfer distance ($\times10^{-3}$), lower values indicate better performance.}}
\vspace{-0.5em}
\resizebox{170mm}{!}{
\begin{tabular}{p{2.7cm}<{\centering}|p{1cm}<{\centering}p{1cm}<{\centering}p{1cm}<{\centering}p{1.4cm}<{\centering}p{1cm}<{\centering}p{1cm}<{\centering}p{1cm}<{\centering}p{1cm}<{\centering}p{1cm}<{\centering}p{1cm}<{\centering}|p{1cm}<{\centering}}
\toprule[0.25mm]
{Model}& Bottle & Bowl& Can&Conditioner&Cube& Fork&Gear  &Pepsi&Spoon&Scissors&Average  \\

\midrule[0.15mm]
Chart-based~\cite{DBLP:conf/nips/SmithCRGMMD20}&7.02& \textbf{6.50}&5.27&4.31&5.65&\textbf{4.79}&7.45&5.77&4.95&6.62&5.83\\

\textbf{Ours} &\textbf{6.23} & {7.41}&\textbf{4.62}&\textbf{3.57}&\textbf{4.71}&5.04&\textbf{7.13}&{4.79}&{3.86}&{5.82}&\textbf{5.32}\\

\midrule[0.15mm]
W/o GAP&8.75& 11.24&8.08&6.30&6.29&6.75&10.74&6.16&9.95&6.47&8.07\\

W/o tactile direction&8.12&9.54&6.79&5.01&6.04&6.10&8.61&6.22&4.90&6.30&6.76 \\

$1\times1$ per-pixel fusion&7.55& 8.18&6.34&4.97&6.18&4.92&7.51&5.77&\textbf{3.67}&\textbf{5.53}&6.06\\

$5\times5$ per-pixel fusion&6.80& {7.10}&5.66&5.73&5.36&5.17&8.73&\textbf{4.65}&5.36&6.69&6.13\\

Single-stage fusion&6.75& 8.06&5.38&4.20&5.29&4.91&8.10&5.75&4.74&6.20&5.94\\

\bottomrule[0.25mm]
\end{tabular}}
\label{tabs:detialed_compare}
\end{table*}

\begin{figure}[t!]
\centering
\includegraphics[scale=0.242]{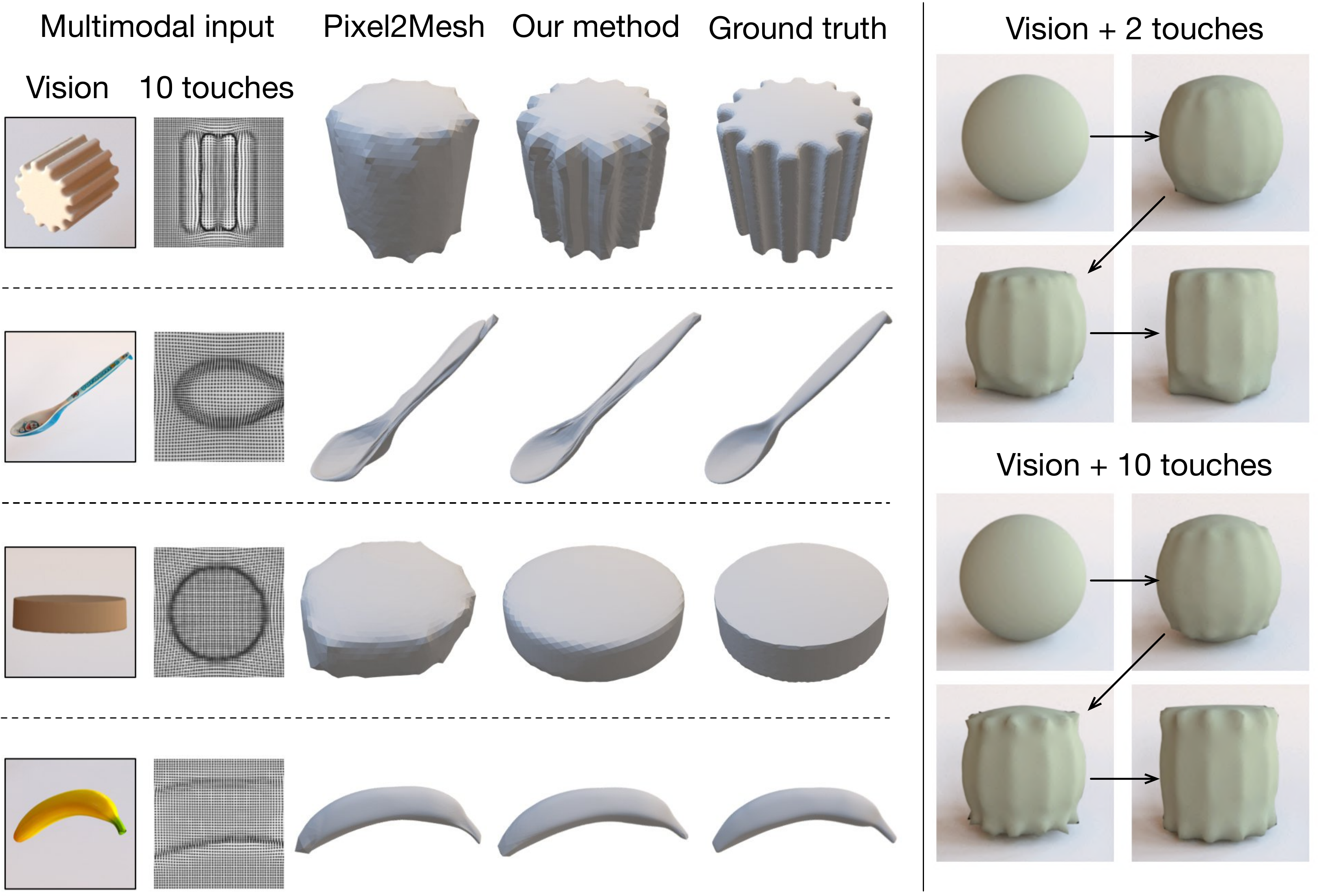}
\caption{\textmd{Left: Visualization of the mesh reconstruction in the test datasets. Our multimodal method achieves  better performance than using the single-view image only, \emph{i.e.} Pixel2Mesh. Note that we use 10  tactile patterns in total from different directions for reconstructing each mesh, and we depict one of them given the limited space. Right: Comparison of deformation processes of using 2 or 10 touches. }}
\label{pic:tactile2mesh}
\vskip -0.15 in

\end{figure}

\subsection{Tactile-Visual Mesh Reconstruction}
\label{sec:exp_mesh}
In this part, we assess the performance of 3D mesh prediction given a single-view image and  a certain number of simulated tactile data by our method. The evaluation is accomplished on 3D mesh models of 10 classes. We collect visual images under random view and tactile data from different press directions. Per each training iteration, we randomly sample 1 visual image and 1$\sim$10 tactile images as the network input. We perform random deformation on meshes so that there are no intersections of 3D meshes for training and testing.

We adopt ResNet-18 as our tactile-visual perception backbone and resize both visual and tactile images to the resolution $224\times224$. In addition, we adopt the cosine learning rate scheduler with an initial learning rate $2\times10^{-5}$, and we train 50 epochs in total.

The quantitative comparison is provided in Table~\ref{tabs:chamfer}, where we calculate the average Chamfer distance as the evaluation metric by sampling both 1,000 points from the predicted and target meshes. We vary the input number of tactile images from 2 to 10 in analogy to the perception task before. Even surprisingly, our method with only 2 touches is sufficient to gain a smaller reconstruction error than Pixel2Mesh (the Visual column in the table), and the error will become much smaller if we increase the number to 10. 
Table~\ref{tabs:chamfer} also provides comparison results with a current SOTA the Chart-based method~\cite{DBLP:conf/nips/SmithCRGMMD20} which adopts estimated local depth and point cloud for tactile-visual fusion. We use the public codes by the authors and share their default setting of generating 5 touches at each time. Thus, the tactile input by~\cite{DBLP:conf/nips/SmithCRGMMD20} could only be the multiplier of 5 (5, 10, etc). According to the results in Table~\ref{tabs:chamfer}, our method surpasses the current the Chart-based method with a large margin, probably thanks to the more elaborate multimodal fusion in our method.

Table~\ref{tabs:detialed_compare} provides detailed evaluation results for each of the 10 classes. We also perform ablation studies to verify the effectiveness of each component that we propose. The ablation variants include: 
\begin{itemize}
    \item W/o GAP: the global average pooling is removed from our model and the feature maps of tactile and visual modalities are fused pixel-by-pixel. This ablation study is to justify the validity of our global-to-local fusion by  GAP on alleviating multimodal  feature misalignment. 
    \item  W/o tactile direction: the learned embedding of tactile direction (as illustrated in Figure~\ref{pic:agg}) is no longer integrated with the Agg embedding.
    \item  $1\times1$ per-pixel fusion: as illustrated in Figure~\ref{pic:agg}, each aggregated tactile embedding is added to the $3\times3$ visual pixels. We change $3\times3$ to 1 single visual pixel to observe if attending the local region is necessary. 
    \item $5\times5$ per-pixel fusion: conversely, we change $3\times3$ to $5\times5$ to see whether the performance is benefited from a larger visual receptive field.
    \item Single-stage fusion: we remove the dense multi-scale connections between the tactile and visual branches and only allow one connection between each corresponding stage. 
    \end{itemize}
    
Observed from Table~\ref{tabs:detialed_compare}, we have the following findings:
\begin{itemize}
    \item GAP plays an essential role in our tactile-visual perception method, as removing GAP leads to noticeable performance drops regarding all objects especially for those elaborate ones (\emph{e.g.} spoon). This is possibly because adopting GAP helps alleviate the misalignment issue of  multimodal features as mentioned in \textsection~\ref{sec:application}.
        \item The tactile direction is also crucial in replenishing the information in tactile embeddings, since the generation of tactile data depends on what touch direction we apply.  
    \item For some elaborate objects, \emph{e.g.} spoon and scissors, adopting $1\times1$ per-pixel fusion can achieve promising or even better performance than the default $3\times3$ setting; similarly, $5\times5$ yields the superiority accuracy regarding some uniform objects. Overall, the $3\times3$ setting comes as the best choice for general classes of the evaluated objects.
    \item Performing single-stage fusion still produces desired results, but it is inferior to the multi-scale version.
\end{itemize} 

Figure \ref{pic:tactile2mesh} displays the generation process of the 3D mesh models. We compare our method with Pixel2Mesh~\cite{wang2018pixel2mesh} that only adopts the visual input. Qualitatively, by adding the tactile input, our approach obtains better predictions than Pixel2Mesh in Figure~\ref{pic:tactile2mesh}. The experimental results here well verify the power of our tactile simulation in capturing the fine-grained patterns of the touched object. 

\subsection{Robot Environment Integration}
\begin{figure}[h!]
\centering
\hskip 0.1 in
\includegraphics[scale=0.25]{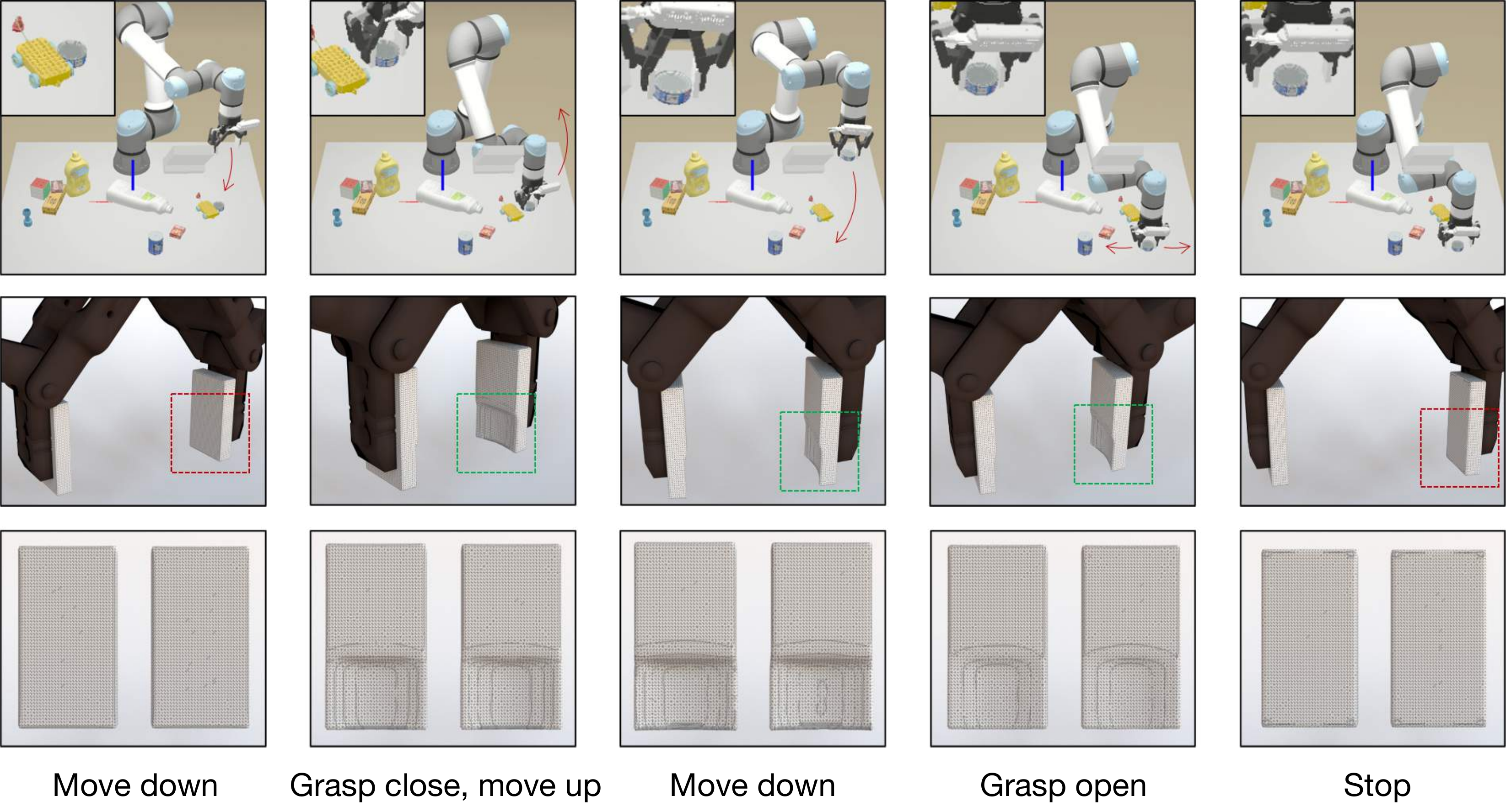}
\caption{\textmd{Illustration of a simulation scene where a robot  picks up a fish can and puts it down. Red/green frames indicate without/with tactile deformation respectively. Zoom in for the best view.}}
\label{pic:robot-can}
\end{figure}

We integrate our tactile simulation with the robot environment, to perform the pick-and-place task for several different objects. 
We first fuse RGB and depth information to get the corresponding semantic segmentation based on the multimodal fusion method in \cite{wang2020cen}. With the segmentation at hand, we detect the 3D position of the can and then pick it up and finally put it down at a different place. The whole process is depicted in Figure \ref{pic:robot-can}, below which we plot the corresponding tactile simulation for each phase. We observe that our tactile simulation does encode the cylinder shape of the can. Besides, the last column  shows that the simulated tactile sensor can return to its original state after the grasping process.

\section{Conclusion}
\vspace{0.03 in}
In this work, we propose Elastic Interaction of Particles (EIP), a new method to simulate interactions between the tactile sensor and the object during robot manipulation. Different from existing tactile simulation methods, our design is based on the elastic interaction of particles, which allows much more accurate simulation with high resolution. Based on our tactile simulation, we further propose a global-to-local tactile-vision perception method for 3D geometric mesh reconstruction. Detailed experimental results verify the effectiveness of our tactile simulation and the proposed tactile-visual perception  scheme.

\vspace{0.03 in}
\section*{Acknowledgement}
\vspace{0.03 in}
This research is jointly funded by Major Project of the New Generation of Artificial Intelligence, China (No. 2018AAA0102900), the Sino-German  Collaborative Research Project Crossmodal Learning (NSFC  62061136001/DFG TRR169), and sponsored by CAAI-Huawei MindSpore Open Fund.

\newpage
\bibliographystyle{splncs04}
\balance
\bibliography{cite}

\end{document}

%% file: math_commands.tex
\usepackage{amsmath,amsfonts,bm}

\def\eqref#1{equation~\ref{#1}}

\def\1{\bm{1}}

\def\vf{{\bm{f}}}
\def\vg{{\bm{g}}}

\def\vv{{\bm{v}}}

\def\vx{{\bm{x}}}

\def\mC{{\bm{C}}}

\def\mF{{\bm{F}}}

\def\mI{{\bm{I}}}

\def\mP{{\bm{P}}}

\def\mR{{\bm{R}}}
\def\mS{{\bm{S}}}

\def\mW{{\bm{W}}}
\def\mX{{\bm{X}}}

\def\mZ{{\bm{Z}}}

\DeclareMathAlphabet{\mathsfit}{\encodingdefault}{\sfdefault}{m}{sl}
\SetMathAlphabet{\mathsfit}{bold}{\encodingdefault}{\sfdefault}{bx}{n}

\def\sG{{\mathbb{G}}}

\def\sP{{\mathbb{P}}}

\def\sS{{\mathbb{S}}}

\newcommand{\R}{\mathbb{R}}

